\documentclass{Interspeech2024}
\usepackage{subcaption}
\usepackage{float}
\usepackage[T1]{fontenc}

\interspeechcameraready

\title{Towards measuring fairness in speech recognition: Fair-Speech dataset}

\name{Irina-Elena}{Veliche}
\name{Zhuangqun}{Huang}
\name{Vineeth}{Ayyat Kochaniyan}
\name{Fuchun}{Peng}
\name{Ozlem}{Kalinli}
\name{Michael L.}{Seltzer}

\address{Meta AI, USA}
\email{ive@meta.com, teddyhuang@meta.com, vineeth.ayyat@gmail.com, fuchunpeng@meta.com, okalinli@meta.com, mikeseltzer@meta.com}


\begin{document}

\maketitle

\begin{abstract}
    
    The current public datasets for speech recognition (ASR) tend not to focus specifically on the fairness aspect, such as performance across different demographic groups. This paper introduces a novel dataset, Fair-Speech, a publicly released corpus to help researchers evaluate their ASR models for accuracy across a diverse set of self-reported demographic information, such as age, gender, ethnicity, geographic variation and whether the participants consider themselves native English speakers. Our dataset includes approximately 26.5K utterances in recorded speech by 593 people in the United States, who were paid to record and submit audios of themselves saying voice commands. We also provide ASR baselines, including on models trained on transcribed and untranscribed social media videos and open source models.
\end{abstract}

\section{Introduction}
The performance of current speech recognition (ASR) systems has improved significantly over the last few years, with the emergence of new modeling techniques and considerable amounts of training data. However, most of the improvements are targeted for overall word error rate (WER). The evaluation sets being used tend to lack information associated with the demographic characteristics, such as ethnicity, geographic variation and whether utterances come from native or non-native English speakers. Also, most numbers are reported in aggregate, without giving a more clear picture of the potential gaps between different demographic groups. While there have been many studies showing that ASR systems do not perform equally well for all demographic and accent groups \cite{gender_dialect_bias, survey_bias_fairness_ml, amazon_fairness2022, racial_disparities_ASR, asr4real, quantifying_bias}, the number of open sourced datasets that can be used for evaluation of such characteristics is limited.

In this paper we introduce a new ASR dataset, Fair-Speech. Our dataset includes approximately 26.5K utterances in recorded speech by 593 people in the U.S. who were paid to record and submit audio of themselves saying commands. They self-identified their demographic information, such as age, gender, ethnicity, geographic location and whether they consider themselves native English speakers, together with their first language.

The verbal commands included in this dataset are categorized into seven domains, primarily serving voice assistant use cases — music, capture, utilities, notification control, messaging, calling, and dictation — that can support researchers who are building or have models in those areas. In response to prompts that relate to each of these domains, dataset participants provided their own audio commands. Some examples of prompts were asking how they would search for a song or make plans with friends, including deciding where to meet. Providing broad prompts to guide the speakers is better than simply asking participants to read text prompts, since that tends to make the audios sound less natural: people would make different kinds of pauses than in natural speech and entities might also not be pronounced properly, if the participants are not familiar with them. Our dataset includes the audio and transcription of participants’ utterances, together with their self-identified labels across the different demographic categories. The intent is for this to be used for evaluating the performance of existing ASR models. The data user agreement prevents a user from developing models that predicts the value of those labels, but one may measure the performance of different models as a function of those labels.

By releasing this dataset, we hope to further motivate the AI community to continue improving the fairness of speech recognition models, which will help everyone have a better experience using applications with ASR.

\section{Previous work on ASR Fairness}
As voice recognition systems have become more integrated into daily lives, especially through the use of voice assistants, there has been a considerable amount of research showing that those systems exhibit biases when it comes to the performance of the ASR models. For example, \cite{racial_disparities_ASR} studied the ability of different ASR systems to transcribe structure interviews of black and white speakers, finding that all of them exhibited substantial racial disparities. The study was done on Corpus of regional African American language \cite{coraal}, a collection of socio-linguistic interviews with dozens of black individuals who speak African American Vernacular English, and also on Voices of California \cite{voc}, which is a compilation of interviews recorded in both rural and urban areas of California. Prior studies also showed disparities across accents and socio-economic status of the speakers in \cite{asr4real}, race and gender bias in \cite{race_gender_bias, bias_race, effects_talker_dialect_gender_race, racial_disparities}, regional and non-native accent \cite{quantifying_bias}.

There is also some recent work on how some of these demographic biases can be mitigated, such as in \cite{amazon_fairness2022, veliche2023improving, counterfactually_fair_asr, counterfactual_fairness}. However, some are using in-house datasets, which are difficult to use for comparison. 

There are a number of open sourced datasets that can be used for measure the fairness of ASR systems across different demographic groups. Apart from the two mentioned above \cite{coraal, voc}, there is also the Artie Bias corpus \cite{artie-bias-corpus}, a curated set of the Mozilla Common Voice corpus, which contains demographic tags for age, gender, accent. Casual conversations dataset \cite{casual_conv} has associated tags for gender, age and skin tone, while the ICSI meeting corpus \cite{icsi} has associated information on gender, age, native language and education level.

The Fair-Speech dataset aims to provide data recorded in a free speech manner from a more diverse set of speakers, where participants self-identified across the different demographic categories.

\section{Corpus contents}
The verbal commands included in this dataset are categorized into seven domains, primarily serving voice assistant use cases — music, capture, utilities, notification control, messaging, calling, and dictation. In response to prompts that relate to each of these domains, dataset participants provided their own audio commands. Our dataset includes the audio and transcription of participants’ utterances. The audio is mobile collected. The intention of this dataset is to be used as an evaluation tool, to uncover gaps or biases in ASR models. 

This dataset was constructed with the recordings of paid participants who have explicitly provided their consent for their recordings to be used in research, together with the associated demographic information. This ensures that the dataset aligns with the ethical standards e.g. for data collection, respects the privacy and autonomy of the participants, but also promotes transparency and other key ethical considerations in responsible data collection practices.

Table\ref{tab:fairness_distribution} shows the per-category distribution of the entire dataset, in terms of number of unique speakers and number of utterances for each demographic sub-group. For age we have a fairly good representation across 18 - 65 groups, with a larger percentage of utterances in the 31 - 45 bucket. The 66+ bucket had too few speakers and utterances, so we chose to not include it here. For gender distribution the percentage of utterances is more balanced. Since we didn't have a significant number of utterances from people who identified as non-binary, we chose to not include them, to not show skewed results. In terms of ethnicity, we have a fairly good representation across multiple categories. The two categories where we have less representation are Native Hawaiian or other Pacific Islander, with 3.6\% of total entries and Middle Eastern or North African, with 2.4\%. They also have less number of speakers than the other categories. In terms of geographic variation, a bit more than half of the utterances are from people who earn less than \$50k per year and there are only 7\% from people who earn more than \$100k, with only 50 speakers in that sub-group. In terms of linguistic variation, we split the utterances based on native language, whether it's English or not. There is about 80\% of the data coming from people whose native language is English. For the other bucket, we provide the first language in the dataset as well. After English, most utterances are from Spanish and Mandarin speakers, with other languages represented in smaller percentages as well. 

\begin{table}
\large
 \caption{Fair-speech dataset per-category distribution}
 \label{tab:fairness_distribution}
  \centering
  \resizebox{\columnwidth}{!}{%
  \begin{tabular}{l|c|c}
    \toprule
    \cmidrule(r){2-3}    
    \emph{Age} & \shortstack{\emph{\#speakers}} & \shortstack{\emph{\#utterances}} \\
    \midrule
    \emph{18 - 22} & 84 & 3846 \\
    \emph{23 - 30} & 95 & 4168 \\
    \emph{31 - 45} & 285 & 12770 \\
    \emph{46 - 65} & 129 & 5687 \\
    \bottomrule
    \toprule
    \cmidrule(r){2-3}    
    \emph{Gender}& \shortstack{\emph{\#speakers}} & \shortstack{\emph{\#utterances}} \\
    \midrule
    \emph{Female} & 321 & 14422 \\
    \emph{Male} & 272 & 12049 \\
    \bottomrule
    \toprule
    \cmidrule(r){2-3}    
    \emph{Ethnicity}& \shortstack{\emph{\#speakers}} & \shortstack{\emph{\#utterances}} \\
    \midrule
    \emph{Asian, South Asian or Asian American} & 82 & 3854 \\
    \emph{Black or African American} & 180 & 7807 \\
    \emph{Hispanic, Latino or Spanish} & 63 & 2814 \\
    \emph{Middle Eastern or North African} & 17 & 749 \\
    \emph{Native American, American Indian, or Alaska Native} & 105 & 4632 \\
    \emph{Native Hawaiian or Other Pacific Islander} & 22 & 969 \\
    \emph{White} & 124 & 5646 \\
    \bottomrule
    \toprule
    \cmidrule(r){2-3}    
    \emph{Geographic variation}& 
    \shortstack{\emph{\#speakers}} &
    \shortstack{\emph{\#utterances}} \\
    \midrule
    \emph{Low (US <= \$50k)} & 335 & 14780 \\
    \emph{Medium (US = \$50k-\$100k)} & 217 & 9824 \\
    \emph{Affluent (US = \$100k+)} & 41 & 1867 \\
    \bottomrule
    \toprule
    \cmidrule(r){2-3}    
    \emph{Linguistic variation}& 
    \shortstack{\emph{\#speakers}} &
    \shortstack{\emph{\#utterances}} \\
    \midrule
    \emph{L1} & 482 & 21528 \\
    \emph{L2} & 111 & 4943 \\
    \bottomrule
  \end{tabular}
  }
  \label{tab:data}
\end{table}

\subsection{Breakdown by gender for demographic categories}
\begin{figure*}[h!]
  \begin{subfigure}[b]{0.48\textwidth}
    \includegraphics[width=\textwidth]{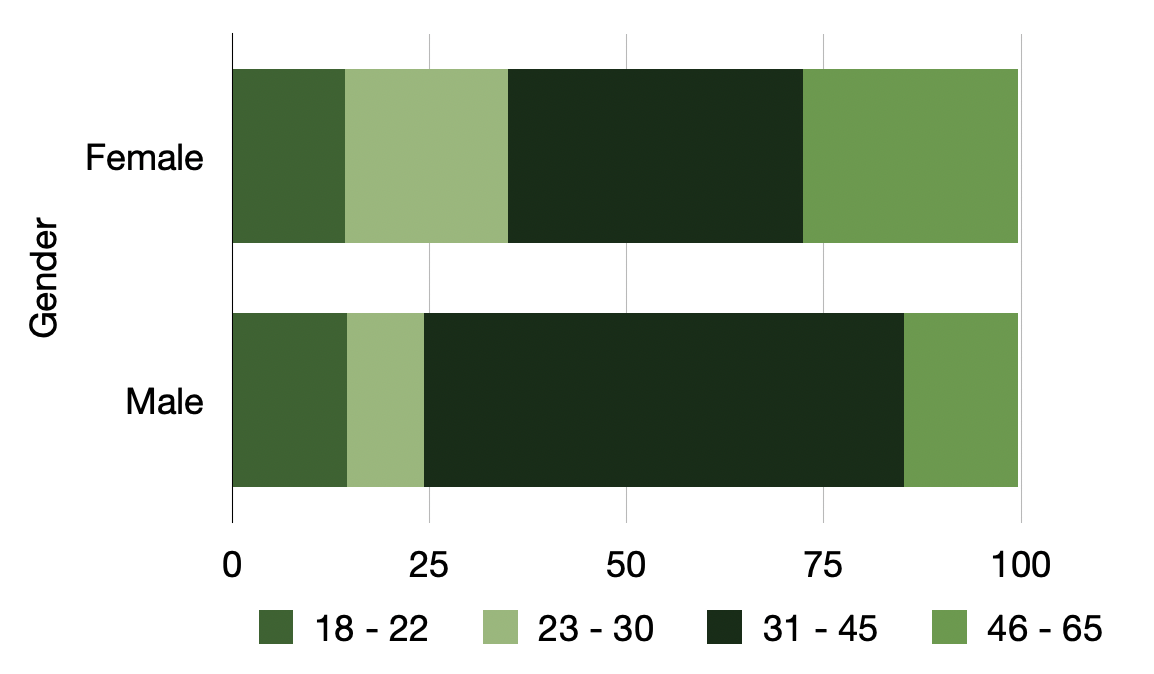}
    \caption{Age breakdown by gender}
    \label{fig:f1}
  \end{subfigure}
  \hfill
  \begin{subfigure}[b]{0.48\textwidth}
    \includegraphics[width=\textwidth]{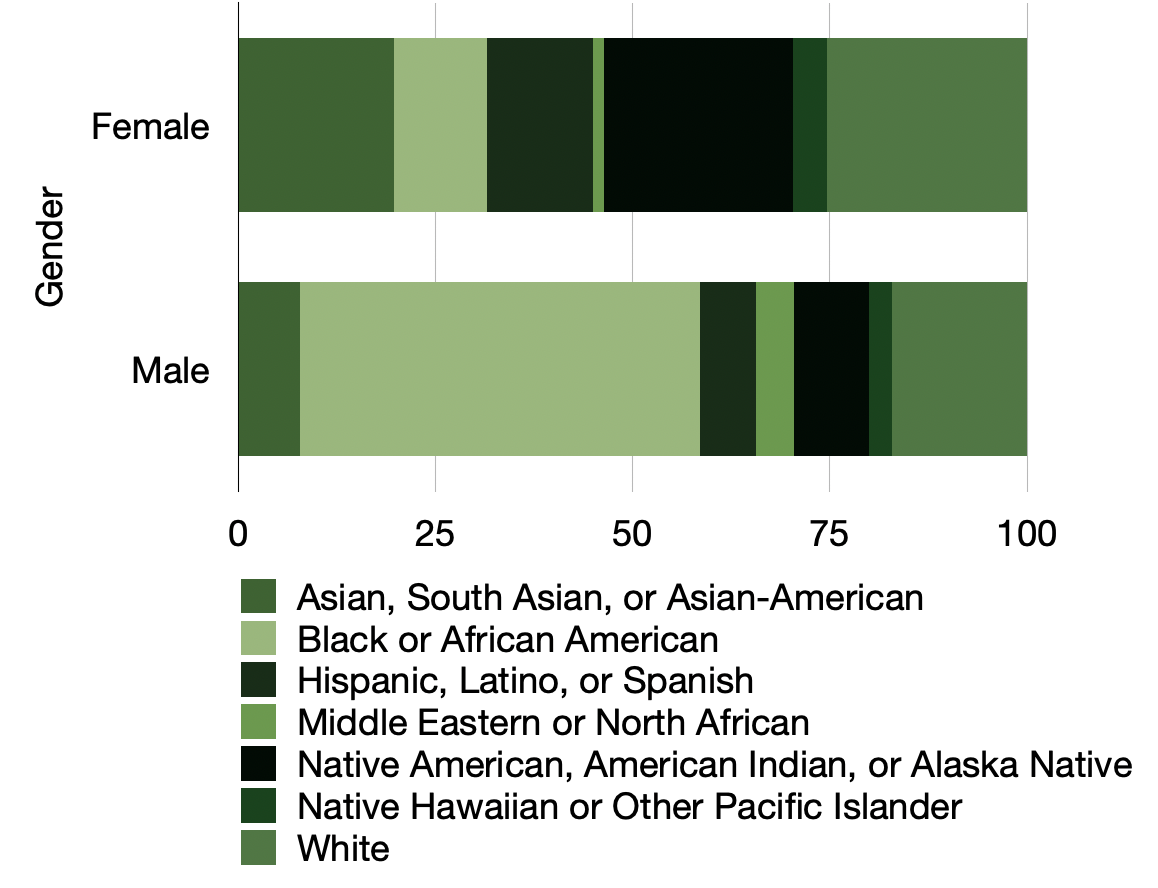}
    \caption{Ethnicity breakdown by gender}
    \label{fig:f1}
  \end{subfigure}
  \hfill
  \begin{subfigure}[b]{0.48\textwidth}
    \includegraphics[width=\textwidth]{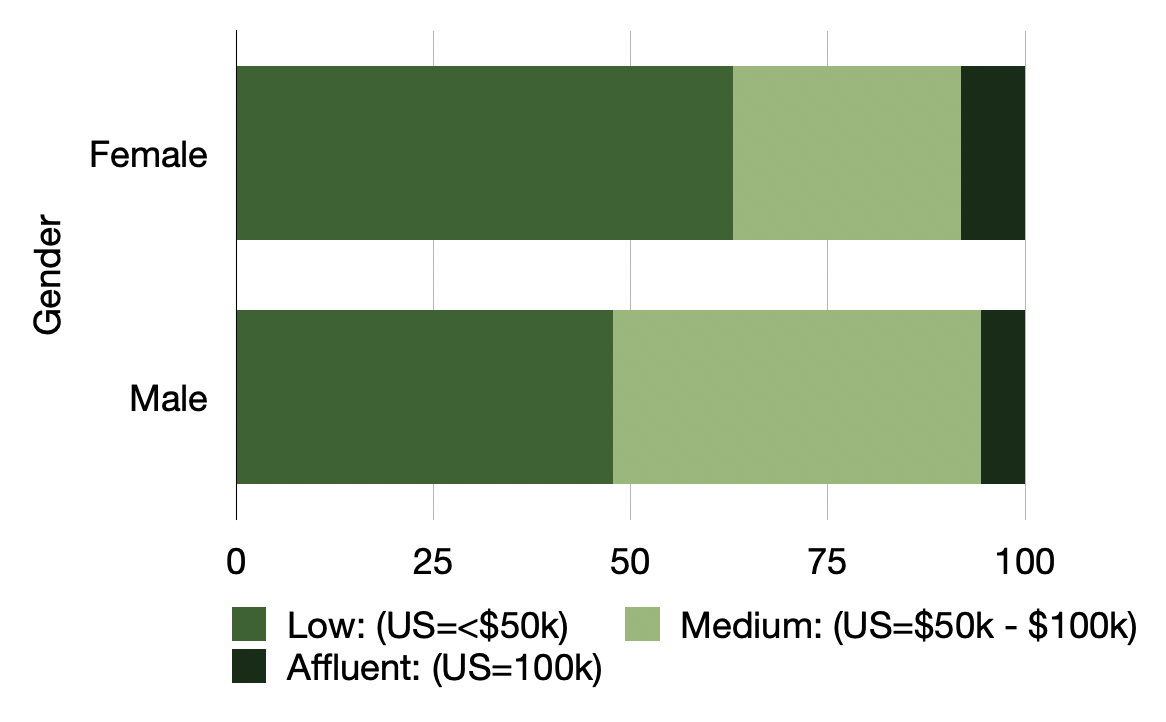}
    \caption{Geographic variation breakdown by gender}
    \label{fig:f1}
  \end{subfigure}
  \hfill
  \begin{subfigure}[b]{0.48\textwidth}
    \includegraphics[width=\textwidth]{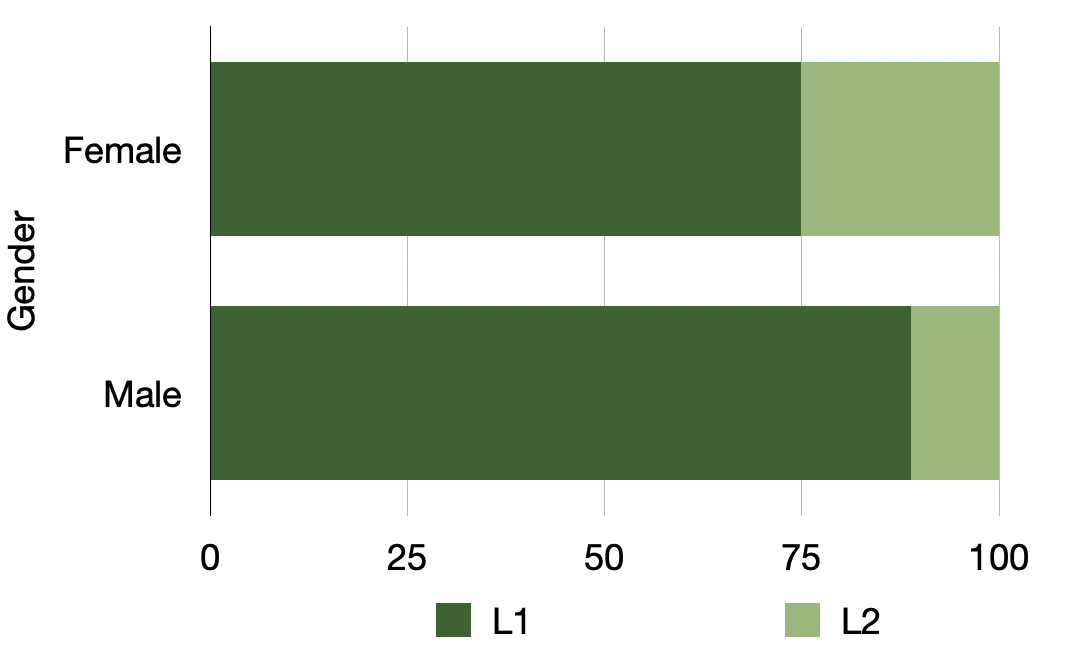}
    \caption{Linguistic variation breakdown by gender}
    \label{fig:f1}
  \end{subfigure}
  \caption{Fairness dataset per-category distribution across demographic groups defined by the gender category.}
  \label{fig:breakdowns}
\end{figure*}

In Figure~\ref{fig:breakdowns} we show the breakdowns by gender for each of the demographic categories. While for some categories there is a good balance between the different genders, for example in the 18 - 22 sub-group, for some there are many more utterances coming from male than from female speakers, such as in the Black or African American sub-group, or the other way around, such as in the Asian, South Asian or Asian-American sub-group. This is important to take into account when analyzing results. Therefore, in section 5.1 we'll take into account confounding factors and speaker variability when showing WER gaps between different sub-groups.

\subsection{Data transcription}
To produce transcription useful for ASR evaluation, all data was verbatim transcribed. Colloquial words were kept as spoken, as well as repeated words. Numbers were spelled out as spoken and entities were transcribed with capital letter. All the other text is lower-cased, without punctuation. Since this is a voice command dataset, each audio contains utterances from a single speaker. The length of the utterances vary. Each recording lasts an average of 7.36 seconds. The maximum length of an utterance is around 1 minute.

High-quality annotations were achieved through a multi-pass human transcription, resolution, curation, and scoring. The multi-pass annotation by external vendors ran until a three-way agreement. The ~10\% without agreement was resolved by internal linguistic engineers. Data curation started from a side by side audit comparing annotations with ASR hypotheses coming from models with different sizes. The ~650 rows with low confidence were further audited by cultural-specific linguistic vendors. All audited data was reviewed by multiple internal teams for cross-functional validation. Finally, 200 random rows were scored by an internal linguistic expert to obtain the annotation WER (1.47\%).

\section{Speech recognition experiments}
To provide some ASR baselines on the ASR fairness dataset on, we trained a series of recurrent neural network transducer (RNN-T) \cite{rnn-t} models and also used open-sourced models. 
\begin{itemize}
    \item 1. Video model, supervised only data: an RNN-T model with an emformer encoder \cite{emformer}, LSTM predictor and a joiner, having approximately 50 million parameters in total. The input feature stride is 6. Encoder network has 13 emformer layers, each with embedded dimension of 480, 4 attention heads, FFN size of 2048. Prediction network is an LSTM layer with 256 hidden units and dropout 0.3. Joint network has 1024 hidden units and a softmax layer of 4096 units for blank and wordpieces. The model was trained on 29.8K hours of English video data that is completely de-identified before transcription. It contains a diverse range of speakers, accents, topics and acoustic conditions. We apply distortion and additive noise to the speed-perturbed data. This results in a total of 148.9K hours of training data.
    \item 2. Video model, semi-supervised: a streaming emformer model trained on over 2 million hours of social media videos. 29.8K hours are manually transcribed as described above and the rest is unlabeled data, decoded by larger teacher models. The model has approximately 290M parameters.
    \item 3. Whisper \cite{whisper}: transformer based model, trained on 1 million hours of weakly labeled audio and 4 million hours of pseudo-labeled audio. We evaluate on both the large-v2 model, which has 1550M parameters and small, which has 244M parameters.
\end{itemize}

\section{Results} \label{sec_results}
\graphicspath{ {./} }

We computed WER on the Fair-Speech dataset using the four models described above. We also noted the relative gap between the groups with the lowest and highest WER in each category. As can be seen in Tables \ref{tab:fairness_eval}, there are gaps across all demographic groups. Some of the key observations:
\begin{itemize}
    \item The data used in training can have a significant impact on the WER, particularly when it comes to bias between different demographic groups. The relative gap across all the demographic categories for all evaluated models is in double digits for most of the categories. For Whisper models the gap is larger than 40\% across all dimensions.
    \item Adding more data in training can significantly improve the performance of the model, such as when semi-supervised data was added for the video models. Interestingly however, for geographic variation, even though the WER improves with more data, the relative gap becomes wider. The linguistic variation difference almost disappears when more data is added in the training, making the dataset more diverse. Also for ethnicity, while the supervised model has some large gaps between different groups, after adding semi-supervised data the gaps decrease and the group with the highest WER actually changes.
    \item As expected, having a larger model improves the accuracy across all the categories, as can be seen with the two Whisper models. However, the relative gap in the linguistic variation sub-group is actually increasing for the larger model.
    \item Data might not be enough to achieve a fair model. All the models shown here were trained on more than 1 million hours of data. However, they exhibit significant gaps across each of the demographic sub-groups. Thus, new modeling techniques are needed to focus on improving the performance for all people. Also, during evaluation, a particular focus needs to be given to demographic breakdowns in addition to overall model accuracy.
\end{itemize} 

\begin{table}
\large
 \caption{Evaluation results on the fairness dataset.}
 \label{tab:fairness_eval}
  \centering
  \resizebox{\columnwidth}{!}{%
  \begin{tabular}{l|c|c|c|c}
    \toprule
    \cmidrule(r){2-5}    
    \emph{Age} & \shortstack{\emph{Video, supervised WER}} & \shortstack{\emph{Video, semi-supervised WER}} &
    \shortstack{\emph{Whisper - small WER}} &
    \shortstack{\emph{Whisper - large-v2 WER}} \\
    \midrule
    \emph{18 - 22} & \textbf{6.52} & \textbf{3.79} & 5.63 & 4.46\\
    \emph{23 - 30} & 7.93 & 4.13 & 6.74 & 4.62 \\
    \emph{31 - 45} & \textbf{11.46} & \textbf{5.16} & \textbf{12.47} & \textbf{7.48} \\
    \emph{46 - 65} & 6.94 & 4.62 & \textbf{5.05} & \textbf{3.65} \\
    \midrule
    \emph{rel. WER gap} & 43.1\% & 26.55\% & 59.5\% & 51.2\%\\
    \bottomrule
    \toprule
    \cmidrule(r){2-5}    
    \emph{Gender}& \shortstack{\emph{Video, supervised WER}} & \shortstack{\emph{Video, semi-supervised WER}} &
    \shortstack{\emph{Whisper - small WER}} &
    \shortstack{\emph{Whisper - large-v2 WER}} \\
    \midrule
    \emph{Female} & \textbf{6.76} & \textbf{3.82}& \textbf{5.16} & \textbf{3.86} \\
    \emph{Male} & \textbf{12.06} & \textbf{5.75}& \textbf{13.3} & \textbf{7.91} \\
    \midrule
    \emph{rel. WER gap} & 43.94\% & 33.56\% & 61.2\% & 51.2\%\\
    \bottomrule
    \toprule
    \cmidrule(r){2-5}    
    \emph{Ethnicity}& \shortstack{\emph{Video, supervised WER}} &  \shortstack{\emph{Video, semi-supervised WER}} &
    \shortstack{\emph{Whisper - small WER}} &
    \shortstack{\emph{Whisper - large-v2 WER}}  \\
    \midrule
    \emph{Asian, South Asian or Asian American} & 6.75 & 4.21 & 4.93 & \textbf{3.7} \\
    \emph{Black or African American} & \textbf{14.21} & 4.9 & \textbf{16.99} & \textbf{9.52} \\
    \emph{Hispanic, Latino or Spanish} & 7.68 & \textbf{5.09} & 5.84 & 3.9\\
    \emph{Middle Eastern or North African} & 8.13 & \textbf{3.67} & 7.99 & 5.08 \\
    \emph{Native American, American Indian, or Alaska Native} & 7.15 & 4.13 & 5.3 & 4.12 \\
    \emph{Native Hawaiian or Other Pacific Islander} & 6.47 & 3.51 & 5.99 & 3.84 \\
    \emph{White} & \textbf{6.29} & 4.03 & \textbf{4.51} & 3.96 \\
    \midrule
    \emph{rel. WER gap} & 55.73\% & 31.04\% & 73.45\% & 61.13\% \\
    \bottomrule
    \toprule
    \cmidrule(r){2-5}    
    \emph{Geographic variation}& 
     \shortstack{\emph{Video, supervised WER}} &  \shortstack{\emph{Video, semi-supervised WER}} &
     \shortstack{\emph{Whisper - small WER}} &
    \shortstack{\emph{Whisper - large-v2 WER}}  \\
    \midrule
    \emph{Low (US <= \$50k)} & 8.67 & \textbf{4.95} & 7.69 & 5.4 \\
    \emph{Medium (US = \$50k-\$100k)} & \textbf{10.13} & 4.54 & \textbf{10.94} & \textbf{6.38}\\
    \emph{Affluent (US = \$100k+)} & \textbf{6.99} & \textbf{3.07} & \textbf{5.55} & \textbf{3.62}\\
    \midrule
    \emph{rel. WER gap} & 30.99\% & 37.97\% & 49.26\% & 43.26\%\\
    \bottomrule
    \toprule
    \cmidrule(r){2-5}    
    \emph{Linguistic variation}& 
     \shortstack{\emph{Video, supervised WER}} &  \shortstack{\emph{Video, semi-supervised WER}} &
     \shortstack{\emph{Whisper - small WER}} &
    \shortstack{\emph{Whisper - large-v2 WER}} \\
    \midrule
    \emph{L1} & \textbf{9.51} & \textbf{4.66} & \textbf{9.54} & \textbf{6.11} \\
    \emph{L2} & \textbf{7.49} & \textbf{4.71} & \textbf{5.63} & \textbf{3.78} \\
    \midrule
    \emph{rel. WER gap} & 21.24\% & 1.06\% & 40.98\% & 46.83\% \\
    \bottomrule
  \end{tabular}
  }
  \label{tab:data}
\end{table}

There can be many nuances when interpreting these WER gaps, due to speaker variability, how many samples we have and confounding factors. Thus, we use an model-based approach to measure fairness, that takes into account all these factors and provides a more accurate picture of the statistical significance of the results. 

\subsection{Understanding the WER gaps}

When analyzing the results, we employed a model-based approach to measure fairness, using mixed-effects Poisson regression to interpret any WER differences between subgroups of interest, as described in \cite{zhe_paper}. This helps by taking into account nuisance factors, unobserved heterogeneity across speakers and helps tracing the source of WER gaps between different subgroups. For this analysis, we used the video semi-supervised model. 

We apply the model-based approach, where we fit a mixed-effects Poisson regression with the demographic group we focus on (age, gender etc.) as the fixed effect and speaker label as a random effect. 

When computing the fairness measurement of speech recognition accuracy among different subgroups of the factor $f(\cdot)$, the model is described as follows in \cite{zhe_paper}:
\begin{align}
\label{model_v3_random}
r_i & \overset{\text{i.i.d.}}{\sim} \mathcal{N}(0, \sigma^2) \\
\label{model_v3_iid}
C_{ij} \,| \, \lambda_{ij} & \overset{\text{i.i.d.}}{\sim} Poisson(\lambda_{ij}) \\
\label{model_v3_formula}
\log(\lambda_{ij})& = \log(N_{ij}) + \mu_{f(i)} + r_i + \theta^{T} x_{ij}
\end{align}
where the utterance-level index of subscription notation $ij$ represents the $j$th utterance from the $i$th speaker, $r_i$ denotes the speaker-level random effect that is independently sampled from a Gaussian distribution with mean 0 and variance $\sigma^2$ which is learnable. $\mu_{f(i)}$ is used to denote the fixed effect for the factor $f(\cdot)$ of primary interest, since typically it is at speaker level.

The bootstrap method \cite{bootstrap} is applied to compute the 95\% confidence interval (CI) of the ratio. If the CI does not include the value of one effect, we assume that there is a statistically significant result.

Results are shown in Table~\ref{tab:analyze_res}. For each demographic category, we do pairwise comparison across all subgroups. The rows in bold indicate that the WER diference between two subgroups are statistically significant, while the rest are insignificant.
\begin{itemize}
    \item For age, when we compare the 18 - 22 group, which has the lowest WER, with the other groups, we see statistically significant differences to groups 31 - 45 and 46 - 65, but not to group 23 - 30.
    \item For gender, there is a statistically significant difference between Female and Male speakers, which is in line with sociolinguistic theory, that establishes that women tend to speak in a more standard way than men \cite{gender_diff}.
    \item For ethnicity, when we do the pairwise comparisons, there are statistically significant differences between the Black or African American subgroup and all the other subgroups. This is in line with previous research on racial disparities. \cite{racial_disparities} found that this is due to phonological, phonetic or prosodic characteristics of African American Vernacular English, rather than the grammatical or lexical characteristics.
    \item In terms of geographic variation, there are statistically significant differences between Low and Medium and Medium and Affluent.
    \item For linguistic variation, even though the difference in WER is quite small between the two sub-groups, we see statistically significant differences between people whose first language is English and those who have a different first language.
\end{itemize}

\begin{table}
\large
 \caption{Analyze WER results using model-based approach.}
 \label{tab:analyze_res}
  \centering
  \resizebox{\columnwidth}{!}{%
  \begin{tabular}{l|c|c}
    \toprule
    \cmidrule(r){2-3}    
    \emph{Age}& \shortstack{\emph{WER ratio}} & \shortstack{\emph{Confidence interval}}  \\
    \midrule
    \emph{18 - 22, 23 - 30} & 1.18 & (0.98, 1.41) \\
    \textbf{\emph{18 - 22, 31 - 45}} & 1.43 & (1.22, 1.68) \\
    \textbf{\emph{18 - 22, 46 - 65}} & 1.31 & (1.06, 1.6) \\
    \textbf{\emph{23 - 30, 31 - 45}} & 1.21 & (1.05, 1.4) \\
    \emph{23 - 30, 46 - 65} & 1.1 & (0.92, 1.32) \\
    \emph{31 - 45, 46 - 65} & 0.91 & (0.8, 1.02) \\
    \bottomrule
    \toprule
    \cmidrule(r){2-3}    
    \emph{Gender}& \shortstack{\emph{WER ratio}} & \shortstack{\emph{Confidence interval}} \\
    \midrule
    \textbf{\emph{Female, Male}} & 1.39 & (1.26, 1.53) \\
    \bottomrule
    \toprule
    \cmidrule(r){2-3}    
    \emph{Ethnicity}& \shortstack{\emph{WER ratio}} & \shortstack{\emph{Confidence interval}} \\
    \midrule
    \textbf{\emph{Asian, South Asian or Asian American, Black or African American}} & 1.81 & (1.55, 2.11) \\
    \emph{Asian, South Asian or Asian American, Hispanic, Latino or Spanish} & 0.97 & (0.76, 1.23) \\
    \emph{Asian, South Asian or Asian American, Middle Eastern or North African} & 1.04 & (0.76, 1.43) \\
    \textbf{\emph{Asian, South Asian or Asian American, Native American, American Indian, or Alaska Native}} & 1.35 & (1.1, 1.66) \\
    \emph{Asian, South Asian or Asian American, Native Hawaiian or Other Pacific Islander} & 1.16 & (0.87, 1.54) \\
    \emph{Asian, South Asian or Asian American, White} & 0.92 & (0.76, 1.12) \\
    \textbf{\emph{Black or African American, Hispanic, Latino or Spanish}} & 0.54 & (0.47, 0.62) \\
    \textbf{\emph{Black or African American, Middle Eastern or North African}} & 0.59 & (0.49, 0.7) \\
    \textbf{\emph{Black or African American, Native American, American Indian, or Alaska Native}} & 0.74 & (0.66, 0.83) \\
    \textbf{\emph{Black or African American, Native Hawaiian or Other Pacific Islander}} & 0.63 & (0.54, 0.73) \\
    \textbf{\emph{Black or African American, White}} & 0.51 & (0.45, 0.57) \\
    \emph{Hispanic, Latino or Spanish, Middle Eastern or North African} & 1.07 & (0.76, 1.5) \\
    \textbf{\emph{Hispanic, Latino or Spanish, Native American, American Indian, or Alaska Native}} & 1.38 & (1.1, 1.74) \\
    \emph{Hispanic, Latino or Spanish, Native Hawaiian or Other Pacific Islander} & 1.19 & (0.88, 1.59) \\
    \emph{Hispanic, Latino or Spanish, White} & 0.94 & (0.76, 1.16) \\
    \emph{Middle Eastern or North African, Native American, American Indian, or Alaska Native} & 1.28 & (0.88, 1.88) \\
    \emph{Middle Eastern or North African, Native Hawaiian or Other Pacific Islander} & 1.08 & (0.75, 1.56) \\
    \emph{Middle Eastern or North African, White} & 0.88 & (0.61, 1.25) \\
    \emph{Native American, American Indian, or Alaska Native, Native Hawaiian or Other Pacific Islander} & 0.85 & (0.67, 1.08) \\
    \textbf{\emph{Native American, American Indian, or Alaska Native, White}} & 0.68 & (0.58, 0.8) \\
    \emph{Native Hawaiian or Other Pacific Islander, White} & 0.8 & (0.59, 1.08) \\
    \bottomrule
    \toprule
    \cmidrule(r){2-3}    
    \emph{Geographic variation}& \shortstack{\emph{WER ratio}} & \shortstack{\emph{Confidence interval}} \\
    \midrule
    \textbf{\emph{Low (US <= \$50k), Medium (US = \$50k-\$100k)}} & 1.2 & (1.08, 1.33) \\
    \emph{Low (US <= \$50k), Affluent (US = \$100k+)} & 1.14 & (0.88, 1.48) \\
    \textbf{\emph{Medium (US = \$50k-\$100k), Affluent (US = \$100k+)}} & 0.72 & (0.62, 0.83) \\
    \bottomrule
    \toprule
    \cmidrule(r){2-3}    
    \emph{Linguistic variation}& \shortstack{\emph{WER ratio}} & \shortstack{\emph{Confidence interval}} \\
    \midrule
    \textbf{\emph{L1, L2}} & 0.82 & (0.71, 0.94) \\
    \bottomrule
  \end{tabular}
  }
  \label{tab:data}
\end{table}

\section{Conclusion}
 In this paper, we introduced a new ASR dataset, Fair-Speech, that has metadata attached for different demographic groups (age, gender, ethnicity, geographic and linguistic variation) and can be used for fairness evaluation when developing speech recognition models. We also run baseline analysis on different models and found that there are statistically significant gaps across the different sub-groups for each demographic category. \\
 The datasets, with transcripts and metadata, are all released to the external community. We hope that they will help in evaluating and improving the fairness of speech recognition models.


\clearpage

\bibliographystyle{IEEEtran}
\bibliography{mybib}

\end{document}